\documentclass[11pt,a4paper]{article}
\usepackage{authblk}
\usepackage[hyperref]{emnlp-ijcnlp-2019}
\usepackage{times}
\usepackage{latexsym}

\usepackage{amsmath,amsfonts,bm}

\def\eqref#1{equation~\ref{#1}}

\def\1{\bm{1}}

\def\vc{{\bm{c}}}
\def\vd{{\bm{d}}}

\def\vg{{\bm{g}}}

\def\vq{{\bm{q}}}
\def\vr{{\bm{r}}}
\def\vs{{\bm{s}}}

\def\vu{{\bm{u}}}
\def\vv{{\bm{v}}}
\def\vw{{\bm{w}}}
\def\vx{{\bm{x}}}

\def\mA{{\bm{A}}}

\def\mD{{\bm{D}}}

\def\mQ{{\bm{Q}}}
\def\mR{{\bm{R}}}
\def\mS{{\bm{S}}}

\def\mU{{\bm{U}}}
\def\mV{{\bm{V}}}

\def\mX{{\bm{X}}}

\DeclareMathAlphabet{\mathsfit}{\encodingdefault}{\sfdefault}{m}{sl}
\SetMathAlphabet{\mathsfit}{bold}{\encodingdefault}{\sfdefault}{bx}{n}

\def\gS{{\mathcal{S}}}

\def\gV{{\mathcal{V}}}

\newcommand{\R}{\mathbb{R}}

\usepackage{hyperref}
\usepackage{url}
\usepackage{enumitem}
\usepackage{amssymb}
\usepackage{amsmath}
\usepackage{cleveref}
\usepackage{booktabs}
\usepackage{bm}
\usepackage{algorithm}
\usepackage{algpseudocode}
\usepackage{graphicx}
\usepackage{nccmath}
\usepackage{caption}
\usepackage{makecell}
\usepackage{subcaption}
\usepackage{makecell}
\usepackage{comment}

\algnewcommand{\Inputs}[1]{%
  \State \textbf{Inputs:}
  \Statex \hspace*{\algorithmicindent}\parbox[t]{.8\linewidth}{\raggedright #1}
}
\algnewcommand{\Outputs}[1]{%
  \State \textbf{Outputs:}
  \Statex \hspace*{\algorithmicindent}\parbox[t]{.8\linewidth}{\raggedright #1}
}

\aclfinalcopy 

\newcommand\blfootnote[1]{%
  \begingroup
  \renewcommand\thefootnote{}\footnote{#1}%
  \addtocounter{footnote}{-1}%
  \endgroup
}

\title{Parameter-free Sentence Embedding via Orthogonal Basis}

\author[1\thanks{~~Most of the work was done during internship at Microsoft}]{Ziyi Yang}
\author[2]{Chenguang Zhu}
\author[3]{Weizhu Chen}
\affil[1]{Department of Mechanical Engineering, Stanford university}
\affil[2]{Microsoft Speech and Dialogue Research Group}
\affil[3]{Microsoft Dynamics 365 AI}
\affil[ ]{\texttt{ziyi.yang@stanford.edu}, \texttt{\{chezhu, wzchen\}@microsoft.com}}

\date{}

\begin{document}
\maketitle
\begin{abstract}
We propose a simple and robust non-parameterized approach for building sentence representations. Inspired by the Gram-Schmidt Process in geometric theory, we build an orthogonal basis of the subspace spanned by a word and its surrounding context in a sentence. We model the semantic meaning of a word in a sentence based on two aspects. One is its relatedness to the word vector subspace already spanned by its contextual words. The other is the word's novel semantic meaning which shall be introduced as a new basis vector perpendicular to this existing subspace. Following this motivation, we develop an innovative method based on orthogonal basis to combine pre-trained word embeddings into sentence representations. This approach requires zero parameters, along with efficient inference performance. We evaluate our approach on 11 downstream NLP tasks. Our model shows superior performance compared with non-parameterized alternatives and it is competitive to other approaches relying on either large amounts of labelled data or prolonged training time.\blfootnote{~~Published as a conference paper at EMNLP 2019 (oral presentation) }
  
\end{abstract}

\section{Introduction}
\label{sec:intro}
The concept of word embeddings has been prevalent in NLP community in recent years, as they can characterize semantic similarity between any pair of words, achieving promising results in a large number of NLP tasks \citep{mikolov2013distributed, pennington2014glove, Salle2016MatrixFU}. However, due to the hierarchical nature of human language, it is not sufficient to comprehend text solely based on isolated understanding of each word. This has prompted a recent rise in search for semantically robust embeddings for longer pieces of text, such as sentences and paragraphs.

Based on learning paradigms, the existing approaches to sentence embeddings can be categorized into two categories: i) parameterized methods and ii) non-parameterized methods. 

\textbf{Parameterized sentence embeddings.} These models are parameterized and require training to optimize their parameters. SkipThought \citep{kiros2015skip} is an encoder-decoder model that predicts adjacent sentences. \citet{pgj2017unsup} proposes an unsupervised model, Sent2Vec, to learn an n-gram feature in a sentence to predict the center word from the surrounding context. Quick thoughts (QT) \citep{logeswaran2018efficient} replaces the encoder with a classifier to predict context sentences from candidate sequences. \citet{khodak2018carte} proposes $\grave{a}\: la \: carte$ to learn a linear mapping to reconstruct the center word from its context. \citet{conneau-EtAl:2017:EMNLP2017} generates the sentence encoder InferSent using Natural Language Inference (NLI) dataset. Universal Sentence Encoder \citep{2018arXiv180407754Y, cer2018universal} utilizes the emerging transformer structure \citep{vaswani2017attention, devlin2018bert} that has been proved powerful in various NLP tasks. The model is first trained on large scale of unsupervised data from Wikipedia and forums, and then trained on the Stanford Natural Language Inference (SNLI) dataset. \citet{wieting-17-recurrent} propose the gated recurrent averaging network (GRAN), which is trained on Paraphrase Database (PPDB) and English Wikipedia. \citet{subramanian2018learning} leverages a multi-task learning framework to generate sentence embeddings. \citet{wieting2016iclr} learns the paraphrastic sentence representations as the simple average of updated word embeddings.

\textbf{Non-parameterized sentence embedding.} Recent work \citep{arora2017asimple} shows that, surprisingly, a weighted sum or transformation of word representations can outperform many sophisticated neural network structures in sentence embedding tasks. These methods are parameter-free and require no further training upon pre-trained word vectors. \citet{arora2017asimple} constructs a sentence embedding called SIF as a sum of pre-trained word embeddings, weighted by reverse document frequency. \citet{ethayarajh2018unsupervised} builds upon the random walk model proposed in SIF by setting the probability of word generation inversely related to the angular distance between the word and sentence embeddings. \citet{2018arXiv180301400R} concatenates different power mean word embeddings as a sentence vector in $p$-mean. As these methods do not have a parameterized model, they can be easily adapted to novel text domains with both fast inference speed and high-quality sentence embeddings. In view of this trend, our work aims to further advance the frontier of this group and make its new state-of-the-art.

In this paper, we propose a novel sentence embedding algorithm, Geometric Embedding (GEM), based entirely on the geometric structure of word embedding space. Given a $d$-dim word embedding matrix $\mA\in\R^{d\times n}$ for a sentence with $n$ words, any linear combination of the sentence's word embeddings lies in the subspace spanned by the $n$ word vectors. We analyze the geometric structure of this subspace in $\R^{d}$. When we consider the words in a sentence one-by-one in order, each word may bring in a novel orthogonal basis to the existing subspace. This new basis can be considered as the new semantic meaning brought in by this word, while the length of projection in this direction can indicate the intensity of this new meaning. It follows that a word with a strong intensity should have a larger influence in the sentence's meaning. Thus, these intensities can be converted into weights to linearly combine all word embeddings to obtain the sentence embedding. In this paper, we theoretically frame the above approach in a QR factorization of the word embedding matrix $\mA$.
Furthermore, since the meaning and importance of a word largely depends on its close neighborhood, we propose the sliding-window QR factorization method to capture the context of a word and characterize its significance within the context. 

In the last step, we adapt a similar approach as \citet{arora2017asimple} to remove top principal vectors before generating the final sentence embedding. This step is to ensure commonly shared background components, e.g. stop words, do not bias sentence similarity comparison. As we build a new orthogonal basis for each sentence, we propose to have disparate background components for each sentence. This motivates us to put forward a sentence-specific principal vector removal method, leading to better empirical results.


We evaluate our algorithm on 11 NLP tasks. Our algorithm outperforms all non-parameterized methods and many parameterized approaches in 10 tasks. Compared to SIF \citep{arora2017asimple}, the performance is boosted by 5.5\% on STS benchmark dataset, and by 2.5\% on SST dataset. Plus, the running time of our model compares favorably with existing models. 

The rest of this paper is organized as following. In \Cref{Aprc}, we describe our sentence embedding algorithm GEM. We evaluate our model on various tasks in \Cref{exp} and \Cref{discussion}. Finally, we summarize our work in \Cref{summary}. Our implementation is available online\footnote{https://github.com/ziyi-yang/GEM}. 

\section{Approach}
\label{Aprc}
We introduce three scores to quantify the importance of a word, as will be explained in this section. First, \textbf{novelty score} $\alpha_n$ measures the portion of the new semantic meaning in a word. Second, \textbf{significance score} $\alpha_s$ describes the alignment between the new semantic meaning and the sentence-level meaning. Finally, \textbf{uniqueness score} $\alpha_u$ examines the uniqueness of the new semantic meaning in the corpus level.
\subsection{Quantify New Semantic Meaning}
\label{sec:newq}
Let us consider the idea of word embeddings \citep{mikolov2013distributed}, where a word $w_i$ is projected as a vector $\vv_{w_i}\in\R^d$. Any sequence of words can be viewed as a subspace in $\R^d$ spanned by its word vectors. Before the appearance of the $i$th word, $\mS$ is a subspace in $\R^d$ spanned by \{$\vv_{w_1}, \vv_{w_2}, ..., \vv_{w_{i-1}}$\}. Its orthonormal basis is $\{\vq_1, \vq_2, ..., \vq_{i-1}\}$. The embedding $\vv_{w_i}$ of the $i$th word $w_i$ can be decomposed into
\begin{equation}
\label{eq: gs}
\begin{aligned}
\vv_{w_i}& = \sum_{j=1}^{i-1}r_j\vq_j + r_{i}\vq_i \\
r_j& = \vq_j^T\vv_{w_i}\\
r_i& = \| \vv_{w_i} - \sum_{j=1}^{i-1}r_j\vq_j\|_2\\ 
\end{aligned}
\end{equation}
where $\sum_{j=1}^{i-1}r_j\vq_j$ is the part in $\vv_{w_i}$ that resides in subspace $\mS$, and $\vq_i$ is orthogonal to $\mS$ and is to be added to $\mS$. The above algorithm is also known as \textbf{Gram-Schmidt Process}.  In the case of rank deficiency, i.e., $\vv_{w_i}$ is already a linear combination of \{$\vq_1, \vq_2, ... \vq_{i-1}$\}, $\vq_i$ is a zero vector and $r_i = 0$. In matrix form, this process is also known as \textit{QR factorization}, defined as follows.


\noindent\textbf{QR factorization.} Define an embedding matrix of $n$ words as $\mA = [\mA_{:, 1}, \mA_{:, 2}, ..., \mA_{:, n}] \in \R^{d \times n}$, where $\mA_{:, i}$ is the embedding of the $i$th word $w_i$ in a word sequence $(w_{1}, \dotsc, w_{i}, \dotsc, w_{n})$. $\mA \in \R^{d \times n}$ can be factorized into $\mA = \mQ\mR$, where the non-zero columns in $\displaystyle \mQ \in \displaystyle \R^{d \times n}$ are the orthonormal basis, and $\displaystyle \mR \in \displaystyle \R^{n\times n}$ is an upper triangular matrix. 


The process above computes the novel semantic meaning of a word w.r.t all preceding words. 
As the meaning of a word influences and is influenced by its close neighbors, we now calculate the novel orthogonal basis vector $\vq_i$ of each word $w_i$ in its neighborhood, rather than only w.r.t the preceding words. 

\textbf{Definition 1 (Contextual Window Matrix)} 

\textit{Given a word $w_i$,  and its $m$-neighborhood window inside the sentence $(w_{i - m}, \dotsc, w_{i-1}, w_i, w_{i + 1}, \dotsc, w_{i + m})$ , define the \textit{contextual window matrix} of word $w_i$ as:}
\begin{equation}
\mS^i = [\vv_{w_{i - m}}... \vv_{w_{i - 1}}, \vv_{w_{i + 1}} ... \vv_{w_{i + m}}, \vv_{w_i}] 
\end{equation}

Here we shuffle $\vv_{w_i}$ to the end of $\mS^i$ to compute its novel semantic information compared with its context. Now the QR factorization of $\mS^i$ is\begin{equation}
\label{eq:QR}
\mS^i = \mQ^i\mR^i
\end{equation}
Note that $\vq_i$ is the last column of $\mQ^i$, which is also the new orthogonal basis vector to this contextual window matrix. 

Next, in order to generate the embedding for a sentence, we will assign a weight to each of its words. This weight should characterize how much new and important information a word brings to the sentence. The previous process yields the orthogonal basis vector $\vq_i$.
We propose that $\vq_i$ represents the novel semantic meaning brought by word $w_i$. We will now discuss how to quantify i) the novelty of $\vq_i$ to other meanings in $w_i$, ii) the significance of $\vq_i$ to its context, and iii) the corpus-wise uniqueness of $\vq_i$ w.r.t the whole corpus.

\subsection{Novelty}
\label{sec:novelty}
We propose that a word $w_i$ is more important to a sentence if its novel orthogonal basis vector $\vq_i$ is a large component in $\vv_{w_i}$, quantified by the proposed novelty score $\alpha_n$. Let $\vr$ denote the last column of $\mR^i$, and $\vr_{-1}$ denote the last element of $\vr$, $\alpha_n$ is defined as:
\begin{equation}
\alpha_n = \exp(\frac{\|\vq_i\|_2}{\|\vv_{w_i}\|_2}) = \exp(\frac{\vr_{-1}}{\|\vr\|_2})
\label{eq:an}
\end{equation}
 Note that $\|\vq_i\|_2 = \vr_{-1}$ and $\|\vv_{w_i}\|_2 = \|\vr\|_2$. One can show that $\alpha_n$ is the exponential of the normalized distance between $\vv_{w_i}$ and the subspace spanned by its context.


\subsection{Significance}
\label{sec:sig}
The significance of a word is related to how semantically aligned it is to the meaning of its context. To identify principal directions, i.e. meanings, in the contextual window matrix $\mS^i$, we employ \textit{Singular Value Decomposition}.


\noindent\textbf{Singular Value Decomposition.} Given a matrix $\mA \in \R^{d \times n}$, there exists $\mU \in \R^{d \times n}$ with orthogonal columns, diagonal matrix $\bm{\Sigma} = \textrm{diag}(\sigma_1,..., \sigma_n)$, $\sigma_1 \geq \sigma_2 \geq ...\geq \sigma_n \geq 0$, and orthogonal matrix $\mV \in \R^{n \times n}$, such that $\mA = \mU \bm{\Sigma} \mV^T$. 

The columns of $\mU$, $\{\mU_{:, j}\}_{j=1}^n$, are an orthonormal basis of $\mA$'s columns subspace and we propose that they represent a set of semantic meanings from the context. Their corresponding singular values $\{\sigma_j\}_{j=1}^n$, denoted by $\sigma(\mA)$, represent the importance associated with $\{\mU_{:, j}\}_{j=1}^n$. The SVD of $w_i$'s contextual window matrix is $\mS^i = \mU^i \bm{\Sigma}^i \mV^{iT} \in \R^{d \times (2m + 1)}$. It follows that $\vq_i^T\mU^{i}$ is the coordinate of $\vq_i$ in the basis of $\{\mU^i_{:, j}\}_{j=1}^{2m + 1}$. 

Intuitively, a word is more important if its novel semantic meaning has a better alignment with more principal meanings in its contextual window. This can be quantified as $\|\sigma(\mS^i)\odot(\vq_i^T\mU^{i})\|_2$, where $\odot$ denotes element-wise product. Therefore, we define the significance of $w_i$ in its context to be:

\begin{equation}
    \alpha_s = \frac{\|\sigma(\mS^i)\odot(\vq_i^T\mU^{i})\|_2}{2m + 1}
\end{equation}

It turns out $\alpha_s$ can be rewritten as

\begin{equation}
\begin{aligned}
    \alpha_s &= \frac{\|\vq_i^T\mU^{i}\bm{\Sigma}^i\|_2}{2m + 1}  =\frac{\|\vq_i^T\mU^{i}\bm{\Sigma}^i\mV^i\|_2}{2m + 1} \\
    &= \frac{\|\vq_i^T\mS^i\|_2}{2m + 1} = \frac{\vq_i^T\vv_{w_i}}{2m + 1} = \frac{\vr_{-1}}{2m + 1}
\end{aligned}
\label{eq:as}
\end{equation}

and we use the fact that $\mV^i$ is an orthogonal matrix and $\vq_i$ is orthogonal to all but the last column of $\mS^i$, $\vv_{w_i}$. Therefore, $\alpha_s$ is essentially the distance between $\vw_i$ and the context hyper-plane, normalized by the context size.

Although $\alpha_s$ and $\alpha_n$ look alike in mathematics form, they model distinct quantities in word $w_i$ against its contextual window. $\alpha_n$ is a function of $\|\vq_i\|_2$ divided by  $\|\vw_i\|_2$, i.e., the portion of the new semantic meaning in word $w_i$. In contrast, \cref{eq:as} shows that $\alpha_s$ equals $\|\vq_i\|_2$ divided by a constant, namely $\alpha_s$ quantifies the absolute magnitude of the new semantic meaning $\vq_i$. 



\subsection{Corpus-wise Uniqueness}
\label{sec:uniq}
Similar to the idea of inverse document frequency (IDF) \citep{sparck1972statistical}, a word that is commonly present in the corpus is likely to be a stop word, thus its corpus-wise uniqueness is small. In our solution, we compute the principal directions of the corpus and then measure their alignment with the novel orthogonal basis vector $\vq_i$. If there is a high alignment, $w_i$ will be assigned a relatively low corpus-wise uniqueness score, and vice versa.

\subsubsection{Compute Principal Directions of Corpus}

In \citet{arora2017asimple}, given a corpus containing a set of $N$ sentences, an embedding matrix $\mX = [\vx_1, \vx_2, \dotsc, \vx_N] \in \R^{d\times N}$ is generated, where $\vx_i$ is the sentence embedding for the $i$-th sentence in the corpus, computed by SIF algorithm. Then principal vectors of $\mX$ are computed and projections onto the principal vectors are removed from each sentence embedding $\vx_i$.

In contrast to \citet{arora2017asimple}, we do not form the embedding matrix after we obtain the final sentence representation. Instead, we obtain an intermediate coarse-grained embedding matrix $\mX^c=[\vg_1, \dotsc, \vg_N]$ as follows. Suppose the SVD of the sentence matrix of the $i$th sentence is $\mS = [\vv_{w_1}, \dotsc, \vv_{w_n}] =\mU \bm{\Sigma} \mV^{T}$. Then the coarse-grained embedding for the $i$th sentence is defined as:


\begin{equation}
\label{eqn:coarse}
\vg_i = \sum_{j=1}^n f(\sigma_j) \mU_{:, j}
\end{equation}
where $f(\sigma_j)$ is a monotonically increasing function. We then compute the top $K$ principal vectors $\{\vd_1, ..., \vd_K\}$ of $\mX^c$, with singular values $\sigma_1 \geq \sigma_2 \geq...\geq \sigma_K$.

\subsubsection{Uniqueness Score}
\label{section:sdr}
In contrast to \citet{arora2017asimple}, we select different principal vectors of $\mX^c$ for each sentence, as different sentences may have disparate alignments with the corpus. 
For each sentence, $\{\vd_1, ..., \vd_K\}$ are re-ranked in descending order of their correlation with sentence matrix $\mS$. The correlation is defined as:
\begin{equation}
    o_i = \sigma_i\|\mS^{T}\vd_i\|_2, 1\leq i \leq K
\end{equation}

Next, the top $h$ principal vectors after re-ranking based on $o_i$ are selected: $\mD=\{\vd_{t_1}, ..., \vd_{t_h}\}$, with $o_{t_1}\geq o_{t_2} \geq ... \geq o_{t_h}$ and their singular values in $\mX^c$ are $\bm{\sigma}_d  = [\sigma_{t_1}, ..., \sigma_{t_h}]\in \R^{h}$. 

Finally, a word $w_i$ with new semantic meaning vector $\vq_i$ in this sentence will be assigned a corpus-wise uniqueness score:
\begin{equation}
\alpha_u = \exp{(-\|\bm{\sigma}_d\odot(\vq_i^T\mD)\|_2/h)}
\label{eq:au}
\end{equation}
This ensures that common stop words will have their effect diminished since their embeddings are closely aligned with the corpus' principal directions. 

\subsection{Sentence Vector}
 A sentence vector $\vc_s$ is computed as a weighted sum of its word embeddings, where the weights come from three scores: a novelty score ($\alpha_n$), a significance score ($\alpha_s$) and a corpus-wise uniqueness score ($\alpha_u$). 
 \begin{equation}\label{eq:alltogether}
\begin{aligned}
\alpha_i &= \alpha_n + \alpha_s + \alpha_u \\
\vc_s &= \sum_{i}\alpha_i \vv_{w_i} \\
\end{aligned}
\end{equation}

We provide a theoretical explanation of \Cref{eq:alltogether} in Appendix.

 \textbf{Sentence-Dependent Removal of Principal Components.} \citet{arora2017asimple} shows that given a set of sentence vectors, removing projections onto the principal components of the spanned subspace can significantly enhance the performance on semantic similarity task. However, as each sentence may have a different semantic meaning, it could be sub-optimal to remove the same set of principal components from all sentences.
 
\begin{figure*}[!h]
\centering
\includegraphics[width=2\columnwidth]{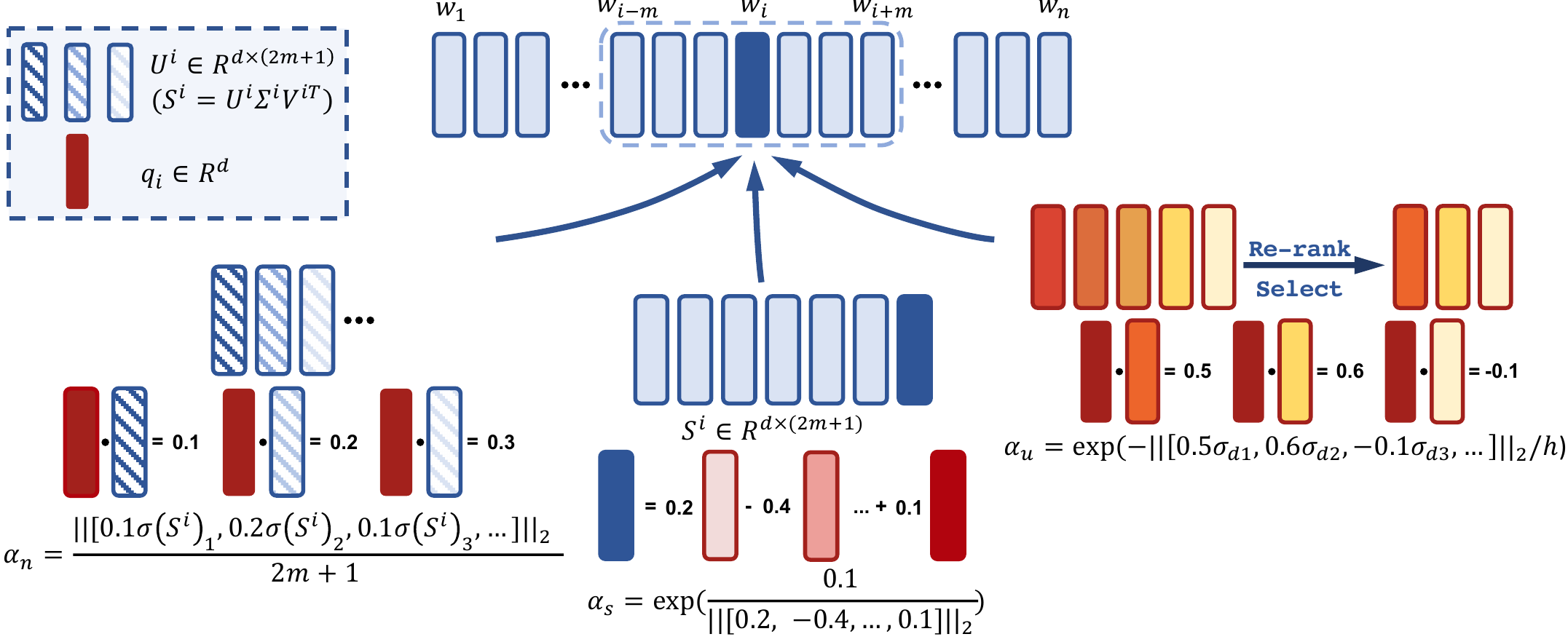}
\caption{An illustration of GEM algorithm. Top middle: The sentence to encode, with words $w_1$ to $w_n$. The contextual window of word $w_i$ is inside the dashed line. Bottom middle: Form contextual window matrix $\mS^i$ for $w^i$, compute $\vq_i$ and novelty score $\alpha_n$ (\Cref{sec:newq} and \Cref{sec:novelty}). Bottom left: SVD of $\mS^i$ and compute the significance score $\alpha_s$ (\Cref{sec:sig}). Bottom right: Re-rank and select from principal components (orange blocks) and compute uniqueness score $\alpha_u$ (\Cref{sec:uniq}).}

\label{fig: algo}
\end{figure*}

 Therefore, we propose the \textit{sentence-dependent principal component removal} (SDR), where we re-rank top principal vectors based on correlation with each sentence. Using the method from \Cref{section:sdr}, we obtain $\mD=\{\vd_{t_1}, ..., \vd_{t_r}\}$ for a sentence $s$. The final embedding of this sentence is then computed as:
 \begin{equation}
     \vc_s \leftarrow \vc_s - \sum_{j=1}^r(\vd_{t_j}^T\vc_s)\vd_{t_j}
 \end{equation} 
 Ablation experiments show that sentence-dependent principal component removal can achieve better result.
The complete algorithm is summarized in \Cref{alg: gem} with an illustration in \Cref{fig: algo}.

\subsection{Handling of out-of-vocabulary Words}
In many NLP algorithms, the out-of-vocabulary (OOV) words are projected to a special ``UNK'' token. However, in this way, different OOV words with drastically different meanings will share the same embedding. To fix this problem, we change this projection method by mapping OOVs to pre-trained in-vocabulary words, based on a hash function SHA-256 of its characters. In this way, two different OOV words will almost certainly have different embeddings. In the experiments, we apply this OOV projection technique in both STS-B and CQA tasks.

\begin{algorithm*}[!h]

\caption{Geometric Embedding (GEM)}\label{alg: gem}
\begin{algorithmic}
\Inputs{A set of sentences $\gS$, vocabulary $\gV$, word embeddings $\{\vv_w\in \R^d\,|\,w\in\gV\}$}
\Outputs{Sentence embeddings \{$\vc_s\ \in \R^d\, |\, s\in \gS$\}}
\For{$i$th sentence $s$ in $\gS$}
\State Form matrix $\mS \in \R^{d \times n}$, $\mS_{:, j} = \vv_{w_j}$ and $w_j$ is the $j$th word in $s$\label{alg: pr1}
\State The SVD is $\mS = \mU \bm{\Sigma} \mV^T$
\State The $i$th column of the coarse-grained sentence embedding matrix $\mX^c_{:, i}$ is $\mU(\sigma(\mS))^3$
\EndFor
\State Take first $K$ singular vectors $\{\vd_1, ..., \vd_K\}$ and singular values $\sigma_1 \geq \sigma_2 \geq...\geq \sigma_K$ of $\mX^c$
\For{sentence $s$ in $\gS$}
\State Re-rank $\{\vd_1, ..., \vd_K\}$  in descending order by $o_i = \sigma_i\|\mS^{T}\vd_i\|_2, 1\leq i \leq K$.
\State Select top $h$ principal vectors as $\mD=[\vd_{t_1}, ..., \vd_{t_h}]$, with singular values $\bm{\sigma}_d=[\sigma_{t_1}...., \sigma_{t_h}]$.\label{alg: pr2}
\For{word $w_i$ in $s$}
\State $\mS^i = [\vv_{w_{i - m}},..., \vv_{w_{i - 1}}, \vv_{w_{i + 1}},...,\vv_{w_{i + m}}, \vv_{w_i}]$ is the contextual window matrix of $w_i$.\label{alg: qr1}
\State Do QR decomposition $\mS^i = \mQ^i\mR^i$, let $\vq_i$ and $\vr$ denote the last column of $\mQ^i$ and $\mR^i$
\State $\alpha_n = \exp(\vr_{-1}/\|\vr\|_2), \alpha_s = \vr_{-1}/(2m + 1), \alpha_u = \exp{({-\|\bm{\sigma}_d\odot(\vq_i^T\mD)\|_2/h})}$
\State $\alpha_i = \alpha_n + \alpha_s + \alpha_u$
\EndFor
\State $\vc_s = \sum_{\vv_i \in \vs}\alpha_i\vv_{w_i}$
\State Principal vectors removal: $\vc_s \leftarrow \vc_s - \mD\mD^T\vc_s$
\EndFor
\end{algorithmic}
\end{algorithm*}

\section{Experiments}
\label{exp}
\subsection{Semantic Similarity Tasks: STS Benchmark}

We evaluate our model on the STS Benchmark \citep{cer2017semeval}, a sentence-level semantic similarity dataset. The goal for a model is to predict a similarity score of two sentences given a sentence pair. The evaluation is by the Pearson's coefficient $r$ between human-labeled similarity (0 - 5 points) and predictions.

\noindent\textbf{Experimental settings.} We report two versions of our model, one only using GloVe word vectors (GEM + GloVe), and the other using word vectors concatenated from LexVec, fastText and PSL \citep{wieting2015paraphrase} (GEM + L.F.P). The final similarity score is computed as an inner product of normalized sentence vectors. Since our model is non-parameterized, it does not utilize any information from the dev set when evaluating on the test set and vice versa. Hyper-parameters are chosen at $m=7$, $h = 17$, $K = 45$, and $t=3$ by conducing hyper-parameters search on dev set. Results on the dev and test set are reported in \Cref{table:sts_r}. As shown, on the test set, our model has a $6.4\%$ higher score compared with another non-parameterized model SIF, and $26.4\%$ higher than the baseline of averaging L.F.P word vectors. It also outperforms all parameterized models including GRAN, InferSent, Sent2Vec and Reddit+SNLI. 

\begin{table}
\resizebox{\columnwidth}{!}{
\centering
\begin{tabular}{ccc}
 \toprule
 Non-parameterized models & dev & test \\ \midrule
 \textbf{GEM + L.F.P (ours)} & 83.5 & 78.4 \\
 \textbf{GEM + LexVec (ours)} & 81.9 & 76.5 \\
 SIF \citep{arora2017asimple} & 80.1 & 72.0 \\
 uSIF \citep{ethayarajh2018unsupervised} & \bf{84.2} & 79.5 \\
 LexVec & 58.78 & 50.43\\ 
 L.F.P & 62.4 & 52.0\\ 
 word2vec skipgram & 70.0 & 56.5\\
 Glove & 52.4 & 40.6\\
 ELMo & 64.6 & 55.9\\ \midrule
 Parameterized models \\ \midrule
 PARANMT-50M \citep{wieting2017paranmt} & - & \bf{79.9} \\
 Reddit + SNLI \citep{2018arXiv180407754Y} & 81.4 & 78.2\\
 GRAN \citep{wieting-17-recurrent}& 81.8 & 76.4 \\
 InferSent \citep{conneau-EtAl:2017:EMNLP2017}& 80.1 & 75.8 \\
 Sent2Vec \citep{pgj2017unsup}& 78.7 & 75.5 \\
 Paragram-Phrase \citep{wieting2016iclr}& 73.9 & 73.2 \\
\bottomrule
\end{tabular}}
\caption{Pearson's $r \times 100$ on STSB. Best results are in bold.}
\label{table:sts_r}
\end{table}

\begin{table}
\begin{center}
\begin{tabular}{cc}
 \toprule
 \textbf{GEM + L.F.P (ours)} & \bf{49.11}\\
 Reddit + SNLI tuned & 47.44 \\ \midrule
 KeLP-contrastive1 & 49.00 \\
 SimBow-contrastive2 & 47.87 \\
 SimBow-primary & 47.22 \\
\bottomrule
\end{tabular}
\caption{MAP on CQA subtask B.}
\label{table:cqa}
\end{center}
\end{table}

\subsection{Semantic Similarity Tasks: CQA}
We evaluate our model on subtask B of the SemEval Community Question Answering (CQA) task, another semantic similarity dataset. Given an original question $Q_o$ and a set of the first ten related questions $(Q_1, ..., Q_{10})$ retrieved by a search engine, the model is expected to re-rank the related questions according to their similarity with respect to the original question. Each retrieved question $Q_i$ is labelled ``PerfectMatch'', ``Relevant'' or ``Irrelevant'', with respect to $Q_o$. Mean average precision (MAP) is used as the evaluation measure.


We encode each question text into a unit vector $\vu$. Retrieved questions $\{Q_i\}_{i=1}^{10}$ are ranked according to their cosine similarity with $Q_o$. Results are shown in \Cref{table:cqa}. For comparison, we include results from the best models in 2017 competition: SimBow \citep{charlet2017simbow}, KeLP \citep{filice2017kelp}, and Reddit + SNLI tuned. Note that all three benchmark models require learning on CQA training set, and SimBow and KeLP leverage optional features including usage of comments and user profiles. In comparison, our model only uses the question text without any training. Our model clearly outperforms Reddit + SNLI tuned, SimBow-primary and KeLP model.

\subsection{Supervised tasks}
We further test our model on nine supervised tasks, including seven classification tasks: movie review (MR) \citep{pang2005seeing}, Stanford Sentiment Treebank (SST) \citep{socher2013recursive}, question-type classification (TREC) \citep{voorhees2003overview}, opinion polarity (MPQA) \citep{wiebe2005annotating}, product reviews (CR) \citep{hu2004mining}, subjectivity/objectivity classification (SUBJ) \citep{pang2004sentimental} and paraphrase identification (MRPC) \citep{dolan2004unsupervised}. We also evaluate on two entailment and semantic relatedness tasks: SICK similarity (SICK-R) and the SICK entailment (SICK-E) \citep{marelli2014semeval}. The sentence embeddings generated are fixed and only the downstream task-specific neural structure is learned. For classification tasks, a linear classifier is trained on top, following \citet{kiros2015skip}, and classification accuracy are reported. For relatedness tasks, we follow \citet{tai2015improved} to train a logistic regression to learn the probability distribution of relatedness scores, and we report Pearson's correlation. The four hyper-parameters are chosen the same as those in STS benchmark experiment. For fair comparison, embeddings models are divided into two categories: non-parameterized and parameterized ones, as described in \cref{sec:intro}. Results are shown in \Cref{table:sup}. 

GEM outperforms all other non-parameterized sentence embedding models, including SIF, p-mean \citep{2018arXiv180301400R}, and BOW on GloVe. The consistent superior performance again demonstrates GEM's advantage on weighting scheme. It also compares favorably with most of parameterized models, including $\grave{a}\: la \: carte$ \citep{khodak2018carte}, FastSent \citep{hill2016learning}, InferSent, QT, Sent2Vec, SkipThought-LN (with layer normalization) \citep{kiros2015skip}, SDAE \citep{hill2016learning}, STN \citep{subramanian2018learning} and USE \citep{2018arXiv180407754Y}. Note that sentence representations generated by GEM have much smaller dimension compared to most of benchmark models, and the subsequent neural structure has fewer trainable parameters. This observation suggests that local multi-word level information in sentences has already provided revealing information for sophisticated downstream tasks. The fact that GEM does well on several classification tasks (e.g. TREC and SUBJ) indicates that the proposed weight scheme is able to recognize important words sentences. Also, GEM's competitive performance on sentiment tasks shows that exploiting the geometric structures of two sentence subspaces is semantically informative. 

\begin{table*}[tbp]
\centering
\resizebox{1\textwidth}{!}{\begin{tabular}[c]{cccccccccccccc}
\toprule
  \textbf{Model}    & \textbf{Dim} & \makecell{\textbf{Training} \\ \textbf{time (h)}} & \textbf{MR}  & \textbf{CR}   & \textbf{SUBJ} & \textbf{MPQA} & \textbf{SST}  & \textbf{TREC} &  \textbf{MRPC} & \textbf{SICK-R} & \textbf{SICK-E} \\ \toprule
\multicolumn{10}{l}{\textit{\normalsize{Non-parameterized models}}} \\ \midrule
\textbf{GEM + L.F.P (ours)}       & 900 & 0 & \textbf{79.8} & \textbf{82.5} & \textbf{93.8} & \textbf{89.9} & \textbf{84.7} & \textbf{91.4} & \textbf{75.4/82.9} &  \textbf{86.5} &  \textbf{86.2}  \\ 
\textbf{GEM + GloVe (ours)}      & 300 & 0 & 78.8 & 81.1 & 93.1 & 89.4 & 83.6 & 88.6 & 73.4/82.3 &  86.3 &  85.3  \\
SIF        & 300 &  0 & 77.3 & 78.6 & 90.5 & 87.0 & 82.2 & 78.0 &    -     &  86.0 &  84.6  \\
uSIF        & 300 &  0 & - & - & - & - & 80.7 & - &    -     &  83.8 &  81.1  \\
p-mean  & 3600 & 0 & 78.4 & 80.4 & 93.1 & 88.9 & 83.0 & 90.6 &  -  &   -   &   -   \\
GloVe BOW  & 300 & 0 & 78.7 & 78.5 & 91.6 & 87.6 & 79.8 & 83.6 & 72.1/80.9 &  80.0  &  78.6  \\ \toprule
\multicolumn{10}{l}{\textit{\normalsize{Paraemterized models}}} \\ \midrule
InferSent & 4096 & 24 & 81.1 & 86.3 & 92.4 & 90.2 & 84.6  & 88.2 & 76.2/83.1 &  88.4  &  86.3  \\ 
Sent2Vec  & 700 & 6.5 & 75.8 & 80.3 & 91.1 & 85.9 &  -   & 86.4 & 72.5/80.8 &   -   &  - \\
SkipThought-LN & 4800 & 336 & 79.4 & 83.1 & 93.7 & 89.3 & 82.9 & 88.4 & - & 85.8 & 79.5      \\
FastSent    & 300 & 2 & 70.8 & 78.4 & 88.7 & 80.6 & - & 76.8 & 72.2/80.3 & - & - \\
$\grave{a}\: la \: carte$ & 4800 & N/A & 81.8 & 84.3 & 93.8 & 87.6 & 86.7 & 89.0 & - & - & - \\
SDAE & 2400 & 192 & 74.6 & 78.0 & 90.8 & 86.9 & - & 78.4 & 73.7/80.7 & - & - \\
QT & 4800 & 28 & 82.4 & 86.0 & \textbf{\underline{94.8}} & 90.2 & \textbf{\underline{87.6}} & 92.4 & 76.9/84.0 & 87.4 & - \\
STN & 4096 & 168 & \textbf{\underline{82.5}} & \textbf{\underline{87.7}} & 94.0 & \textbf{\underline{90.9}} & 83.2 & 93.0 & \textbf{\underline{78.6/84.4}} & \textbf{\underline{88.8}} & \textbf{\underline{87.8}} \\
USE & 512 & N/A & 81.36 & 86.08 & 93.66& 87.14 & 86.24 & \textbf{\underline{96.60}} & - & - & - & \\
\bottomrule
\end{tabular}}
\caption{Results on supervised tasks. Sentence embeddings are fixed for downstream supervised tasks. Best results for each task are underlined, best results from models in the same category are in bold. SIF results are extracted from \citet{arora2017asimple} and \citet{2018arXiv180301400R}, and training time is collected from \citet{logeswaran2018efficient}.}
\label{table:sup}
\end{table*}


\section{Discussion}
\label{discussion}

\textbf{Comparison with \citet{arora2017asimple}.} We would like to point out that although sharing the idea of modelling the sentence as the weighted sum of its word vectors, GEM is substantially different from \citet{arora2017asimple}. First, we adopt well-established numerical linear algebra to quantify the semantic meaning and importance of words in the sentences context. And this new approach proves to be effective. Second, the weights in SIF (and uSIF) are calculated from the statistic of vocabularies on very large corpus (wikipedia). In contrast, the weights in GEM are directly computed from the sentences themselves along with the dataset, independent with prior statistical knowledge of language or vocabularies.  Furthermore, the components in GEM's weights are derived from numerical linear algebra \cref{eq:an} to (\ref{eq:au}), while SIF directly includes a hyper-parameter term in its weight scheme, i.e. its smooth term.

\textbf{Robustness and Effectiveness.}
Besides experiments mentioned above, we also test the robustness and effectiveness GEM on several simple but non-trivial examples. These experiments demonstrate that GEM is quite insensitive against the removal of non-important stop words. Also GEM can correctly assign higher weights for words with more significant semantic meanings in the sentence.

We first test the robustness of GEM by removing one non-important stop word in a sentence and computed the similarity between the original sentence and the one after removal. For example:
\begin{itemize}[noitemsep]
  \item original sentence: ``the first tropical system to slam the US this year is expected to make landfall as a hurricane''
  \item remove 7 stop words: ``first tropical system slam US this year expected make landfall hurricane''
\end{itemize}
The cosine similarity between these two sentences given by GEM is 0.954. Even though aggressively removing 7 stop words, GEM still assigns pretty similar embeddings for these two sentences.

We further demonstrate that GEM does assign higher weights to words with more significant semantic meanings. Consider the following sentence: "there are two ducks swimming in the river". Weights assigned by GEM are (sorted from high to low): [ducks: 4.93, river:4.72 , swimming: 4.70, two: 3.87, are: 3.54, there: 3.23, in:3.04, the:2.93]. GEM successfully assigns higher weight to informative words like “ducks” and “river”, and downplay stop words like “the” and “there”. More examples can be found in the Appendix.

\textbf{Ablation Study.} As shown in in \Cref{table:ablation}, every GEM weight ($\alpha_n, \alpha_s,\alpha_u$) and proposed principal components removal methods contribute to the performance.  
As listed on the left, adding GEM weights improves the score by 8.6\% on STS dataset compared with averaging three concatenated word vectors. The sentence-dependent principal component removal (SDR) proposed in GEM improves 1.7\% compared to directly removing the top $h$ corpus principal components (SIR). Using GEM weights and SDR together yields an overall improvement of 21.1\%.
As shown on the right in \Cref{table:ablation}, every weight contributes to the performance of our model. For example, three weights altogether improve the score in SUBJ task by 0.38\% compared with only using $\alpha_n$.

\begin{table}[htb]
\centering
\scalebox{1}{
\begin{tabular}{ccc}
 \toprule
 Configurations & STSB dev & SUBJ\\ \midrule
 Mean of L.F.P & 62.4 & -\\ 
 GEM weights & 71.0 & -\\
 GEM weights + SIR  & 81.8 & -\\
 GEM weights + SDR & 83.5 & -\\ \midrule
 $\alpha_n$ + SDR & 81.6 & 93.42\\
 $\alpha_n, \alpha_s$ + SDR & 81.9 & 93.6\\
 $\alpha_n, \alpha_s, \alpha_u$ + SDR & 83.5 & 93.8\\
\bottomrule
\end{tabular}}
\caption{Comparison of different configurations demonstrates the effectiveness of our model on STSB dev set and SUBJ. SDR stands for sentence-dependent principal component removal in \Cref{section:sdr}. SIR stands for sentence-independent principal component removal, i.e. directly removing top $h$ corpus principal components from the sentence embedding.
}
\label{table:ablation}
\end{table}

\begin{figure}[!h]
\centering
\includegraphics[width=\columnwidth]{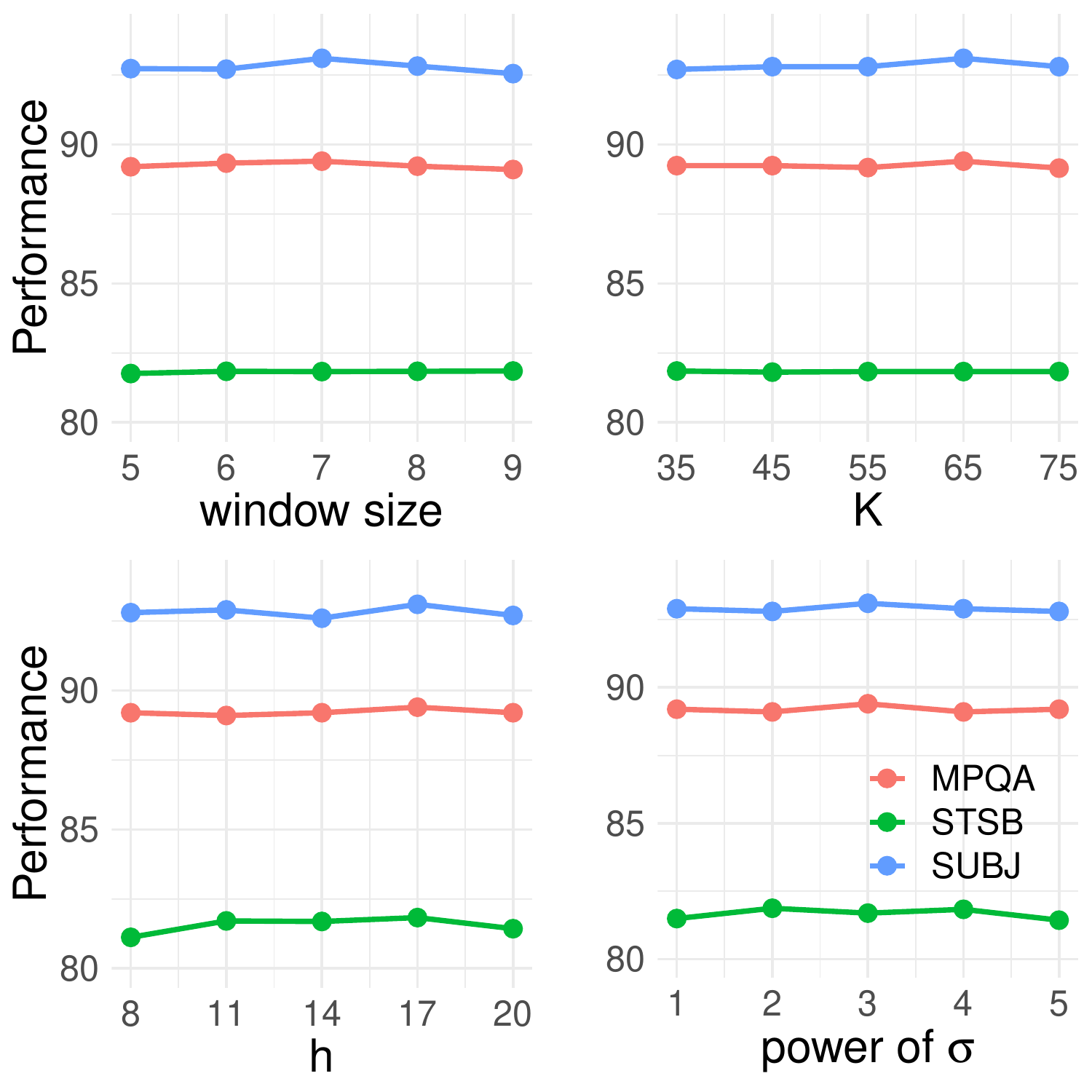}
\caption{Sensitivity tests on four hyper-parameters, the window size $m$ in contextual window matrix, the number of candidate principal components $K$, the number of principal components to remove $h$, and the exponential power of singular value in coarse sentence embedding.}
\label{sensitivity}
\end{figure}

\textbf{Sensitivity Study.} We evaluate the effect of all four hyper-parameters in our model: the window size $m$ in the contextual window matrix, the number of candidate principal components $K$, the number of principal components to remove $h$, and the power of the singular value in coarse sentence embedding, i.e. the power $t$ in $f(\sigma_j) = \sigma_j^t$ in \Cref{eqn:coarse}. We sweep the hyper-parameters and test on STSB dev set, SUBJ, and MPQA. Unspecified parameters are fixed at $m=7$, $K = 45$, $h = 17$ and $t=3$. As shown in \Cref{sensitivity}, our model is quite robust with respect to hyper-parameters.

\textbf{Inference speed.} We also compare the inference speed of our algorithm on the STSB test set with the benchmark models SkipThought and InferSent. SkipThought and InferSent are run on a NVIDIA Tesla P100 GPU, and our model is run on a CPU (Intel Xeon CPU E5-2690 v4 @2.60GHz). For fair comparison, batch size in InferSent and SkipThought is set to be 1. The results are shown in \Cref{table:running_time}. It shows that without acceleration from GPU, our model is still faster than InferSent and is $54\%$ faster than SkipThought.

\begin{table}[htp]
\centering
\scalebox{0.85}{
\begin{tabular}{ccc}
 \toprule
  & Average run time (s) & Variance \\ \midrule
 GEM (CPU) & \textbf{20.08} & 0.23\\
 InferSent(GPU) & 21.24 & 0.15\\
 SkipThought (GPU) & 43.36 & 0.10\\
\bottomrule
\end{tabular}}
\caption{Run time of GEM, InferSent and SkipThought on encoding sentences in STSB test set. GEM is run on CPU, InferSent and SkipThought is run on GPU. Data are collected from 5 trials. \label{table:running_time}}
\end{table}

\section{Conclusions}
\label{summary}
We proposed a simple non-parameterized method to generate sentence embeddings, based entirely on the geometric structure of the subspace spanned by word embeddings. Our sentence embedding evolves from the new orthogonal basis vector brought in by each word, which represents novel semantic meaning. The evaluation shows that our method not only sets up the new state-of-the-art of non-parameterized models but also performs competitively when compared with models requiring either large amount of training data or prolonged training time. In future work, we plan to consider subwords into the model and explore more geometric structures in sentences.

\section*{Acknowledgments}
We would like to thank Jade Huang for proofreading the paper and helpful writing suggestions. We also acknowledge the anonymous reviewers for their valuable feedback.

\bibliography{emnlp-ijcnlp-2019}
\bibliographystyle{acl_natbib}
\clearpage
\appendix
\section{Robustness and Effectiveness of GEM}
\subsection{Robustness Test}
We test the robustness of GEM by removing one non-important stop word in a sentence and computed the similarity between the original sentence and the one after removal. For example:
\begin{itemize}
  \item original = "The student is reading a physics book"
  \item removed = "student is reading a physics book"
\end{itemize}
Stop word "The" is removed. The cosine similarity between embeddings of the two sentences generated by GEM is 0.998. GEM assigns pretty similar embeddings for these two sentences even with the removal of stop words, especially this is a short sentence with only 7 words. More examples are:
\begin{itemize}
  \item original = "Someone is sitting on the blanket"
  \item removed = "Someone is sitting on blanket"
  \item cosine similarity = 0.981
\end{itemize}
and 
\begin{itemize}
  \item original = "A man walks along walkway to the store"
  \item removed = "man walks along walkway to the store"
  \item cosine similarity = 0.984
\end{itemize}
These experiments prove that GEM is robust against stop words and words order.

\subsection{Effectiveness Test}
We also demonstrate that GEM assign higher weights to words with more significant meanings. Consider the sentence: “the stock market closes lower on Friday”, weights assigned by GEM are [lower: 4.94, stock: 4.93, closes: 4.78, market: 4.62, Friday: 4.51, the: 3.75, on: 3.70]. Again, GEM emphasizes informative words like “lower” and “closes”, and diminishes stop words like “the” and “there”.

\section{Proof}

The novelty score ($\alpha_n$), significance score ($\alpha_s$) and corpus-wise uniqueness score ($\alpha_u$) are larger when a word $w$ has relatively rare appearance in the corpus and can bring in new and important semantic meaning to the sentence. 
 
 Following the section 3 in \citet{arora2017asimple}, we can use the probability of a word $w$ emitted from sentence $s$ in a dynamic process to explain \cref{eq:alltogether} and put this as following Theorem with its proof provided below.
 
 
 \textbf{Theorem 1.} \textit{Suppose the probability that word $w_i$ is emitted from sentence $s$ is\footnote{The first term is adapted from \citet{arora2017asimple}, where words near the sentence vector $\vc_s$ has higher probability to be generated. The second term is introduced so that words similar to the context in the sentence or close to common words in the corpus are also likely to occur.}:}
\begin{equation}\label{eq:prob}
    \text{p}[w_i|\vc_s] \propto(\frac{\exp(\langle \vc_s, \vv_{w_i} \rangle)}{Z} + \exp(-(\alpha_n + \alpha_s + \alpha_u)))
\end{equation}
\textit{where $\vc_s$ is the sentence embedding, $Z = \sum_{w_i\in \gV}\exp(\langle \vc_s, \vv_{w_i} \rangle)$ \textit{and} $\gV$ \textit{denotes the vocabulary. }}
\textit{Then when $Z$ is sufficiently large, the MLE for $\vc_s$ is:}
\begin{equation}
\vc_s \propto\sum_{w_i\in s}(\alpha_n + \alpha_s + \alpha_u)\vv_{w_i}
\end{equation}
\\
 \textbf{Proof:}
According to \Cref{eq:prob},
\begin{equation}
    \text{p}[w_i|\vc_s] = \frac{1}{N}(\frac{\exp(\langle \vc_s, \vv_{w_i} \rangle)}{Z} + \exp(-(\alpha_n + \alpha_s + \alpha_u)))
\end{equation}
Where $N$ and $Z$ are two partition functions defined as 
\begin{equation}
\begin{aligned}
    N = 1 + \sum_{w_i\in \gV}&\exp(-(\alpha_n(w_i) + \alpha_s(w_i) + \alpha_u(w_i)))\\
    &Z = \sum_{w_i\in \gV}\exp(\langle \vc_s, \vv_{w_i} \rangle)
\end{aligned}
\end{equation}
The joint probability of sentence $s$ is then
\begin{equation}
p(s|\vc_s) = \prod_{w_i \in s}p(w_i|\vc_s)
\end{equation}
To simplify the notation, let $\alpha = \alpha_n + \alpha_s + \alpha_u$. It follows that the log likelihood $f(w_i)$ of word $w_i$ emitted from sentence $s$ is given by
\begin{equation}
f_{w_i}(\vc_s)  = \log(\frac{\exp(\langle \vc_s, \vv_{w_i} \rangle)}{Z} + e^{-\alpha}) - \log(N)
\end{equation}
\begin{equation}
\nabla f_{w_i}(\vc_s) = \cfrac{\exp(\langle c_s, \vv_{w_i} \rangle)\vv_{w_i}}{\exp(\langle \vc_s, \vv_{w_i} \rangle) + Ze^{-\alpha}}
\end{equation}
By Taylor expansion, we have
\begin{equation}
    \begin{aligned}
    f_{w_i}(\vc_s) &\approx f_{w_i}(0) + \nabla f_{w_i}(0)^T\vc_s \\
    &= \text{constant} + \frac{\langle\vc_s, \vv_{w_i}\rangle}{Ze^{-\alpha} + 1}
    \end{aligned}
\end{equation}
Again by Taylor expansion on $Z$,
\begin{equation}
\begin{aligned}
    \frac{1}{Z e^{-\alpha} + 1} &\approx \frac{1}{1+Z} + \frac{Z}{(1+Z)^2}\alpha \\
    &\approx \frac{Z}{(1+Z)^2}\alpha\\
    &\approx \frac{1}{1+Z}\alpha
\end{aligned}
\end{equation}

The approximation is based on the assumption that $Z$ is sufficiently large. It follows that,
\begin{equation}
    f_{w_i}(\vc_s) \approx \text{constant} + \frac{\alpha}{1+Z}\langle\vc_s, \vv_{w_i}\rangle
\end{equation}

Then the maximum log likelihood estimation of $\vc_s$ is:
\begin{equation}
\begin{aligned}
    \vc_s &\approx \sum_{w_i\in s}\frac{\alpha}{1+Z}\vv_{w_i} \\
    &\propto\sum_{w_i\in s}(\alpha_n + \alpha_s + \alpha_u)\vv_{w_i}
\end{aligned}
\end{equation}

\section{Experimental settings}
For all experiments, sentences are tokenized using the NLTK tokenizer \citep{bird2009natural} wordpunct\_tokenize, and all punctuation is skipped. $f(\sigma_j) = \sigma_j^t$ in \Cref{eqn:coarse}. In the STS benchmark dataset, our hyper-parameters are chosen by conducting parameters search on STSB dev set at $m=7$, $h = 17$, $K = 45$, and $t=3$. And we use the same values for all supervised tasks. The integer interval of parameters search are $m \in [5, 9]$,  $h \in [8, 20]$, $L \in [35, 75]$ (at stride of 5), and $t \in [1, 5]$. In CQA dataset, $m$ and $h$ are changed to 6 and 15, the correlation term in \cref{section:sdr} is changed to $o_i = \|\mS^{T}\vd_i\|_2$ empirically. In supervised tasks, same as \citet{arora2017asimple}, we do not perform principal components in supervised tasks.

\section{Clarifications on Linear Algebra}
\subsection{Encode a long sequence of words}
We would like to give a clarification on encoding a long sequence of words, for example, a paragraph or a article. Specifically, the length $n$ of the sequence is larger than the dimension $d$ of pre-trained word vectors in this case. The only part in GEM relevant to the length of the sequence $n$ is the coarse embedding in \Cref{eqn:coarse}. The SVD of the sentence matrix of the $i$th sentence is still $\mS \in \R^{d\times n} = [\vv_{w_1}, \dotsc, \vv_{w_n}] =\mU \bm{\Sigma} \mV^{T}$, where now $\mU \in \R^{d\times d}$, $\bm{\Sigma}\in \R^{d\times n}$, and $\mV \in R^{n\times n}$. Note that the $d+1$th column to $n$th column in $\bm{\Sigma}$ are all zero. And \Cref{eqn:coarse} becomes $\vg_i = \sum_{j=1}^d f(\sigma_j) \mU_{:, j}$. The rest of the algorithm works as usual. Also, Gram-Schmidt (GS) process is computed in the context window of word $w_i$, and the length of context window is set to be $2m + 1 = 17$ in STS benchmark dataset and supervise downstream tasks. That is, GS is computed on 17 vectors, and 17 is smaller than the dimension $d$. Therefore, GS is always validate in our model, independent with the length of the sentence.

\subsection{Sensitivity to Word Order}
Although utilizing Gram-Schmidt process (GS), GEM is insensitive to the order of words in the sentence, explained as follows. The new semantic meaning vector $q_i$ computed from doing GS on the context window matrix $\mS^{i}$ is independent with the relative order of first $2m$ vectors. This is because in GEM $w_i$ (the word we are calculating weights for) is always shifted to be the last column of $\mS^{i}$. And weighting scheme in GEM only depends on $\vq_i$. Therefore, weight scores stay the same for $w_i$. 

\end{document}